\documentclass[runningheads]{llncs}
\usepackage{graphicx}
\usepackage{comment}
\usepackage{xcolor}
\newcommand{\cy}[1]{}

\begin{document}
\title{My Teacher Thinks The World Is Flat! Interpreting Automatic Essay Scoring Mechanism}
\titlerunning{Interpreting Automatic Essay Scoring Mechanism}
%
\author{Swapnil Parekh\protect\footnote{Equal Contribution}\inst{1} \and Yaman Kumar Singla\protect\footnotemark[2]\inst{1,2,3} \and
 Changyou Chen\inst{3} \and
 Junyi Jessy Li\inst{4} \and
 Rajiv Ratn Shah\inst{1}
}


\institute{IIIT-Delhi \\
\email{swapnilbp100@gmail.com, \{yamank,rajivratn\}@iiitd.ac.in}\and Adobe \\
\email{ykumar@adobe.com} \and State University of New York at Buffalo \\
\email{changyou@buffalo.edu}\and University of Texas at Austin \\ \email{jessy@austin.utexas.edu}}

\authorrunning{S. Parekh et al}
%
\maketitle              

\begin{abstract}
Significant progress has been made in deep-learning based Automatic Essay Scoring (AES) systems in the past two decades. However, little research has been put to understand and interpret the black-box nature of these deep-learning based scoring models. Recent work shows that automated scoring systems are prone to even common-sense adversarial samples. Their lack of natural language understanding capability raises questions on the models being actively used by millions of candidates for life-changing decisions. With scoring being a highly multi-modal task, it becomes imperative for scoring models to be validated and tested on all these modalities. We utilize recent advances in interpretability to find the extent to which features such as coherence, content and relevance are important for automated scoring mechanisms and why they are susceptible to adversarial samples. We find that the systems tested consider essays not as a piece of prose having the characteristics of natural flow of speech and grammatical structure, but as `\emph{word-soups}' where a few words are much more important than the other words. Removing the context surrounding those few important words causes the prose to lose the flow of speech and grammar, however has little impact on the predicted score. We also find that since the models are not semantically grounded with world-knowledge and common sense, adding false facts such as ``the world is flat'' actually increases the score instead of decreasing it. 

\end{abstract}

\section{Introduction}

Automatic Essay Scoring (AES) systems help to alleviate workload of teachers and save time and costs associated with grading. On an average, a British teacher spends 5 hours in a calendar week scoring exams and assignments \cite{micklewright2014teachers,kumar2019get}. This figure is even higher for developing and low-resource countries where the teacher to student ratio is dismal. While on one hand, autograding systems effectively reduce this burden, allowing more working hours for teaching activities, on the other, there have been many complaints against these systems for not scoring the way they are supposed to \cite{flawedAlgos,roboGrade,emptyDream,midDay,perelmanBableWebsite}. For instance, with the recently released Utah Automatic Scoring system, students scored lower by writing question-relevant keywords but higher by including unrelated words and sentences \cite{flawedAlgos,roboGrade}. Similarly, it has been a common complaint that AES systems focus unjustifiably on obscure and difficult vocabulary \cite{perelmanBable}. The concerns are further alleviated by the fact that the scores awarded by these systems are used in life-changing decisions ranging from college and job applications to visa approvals.

Traditionally, autograding systems are built using manually crafted features used with machine learning based models. Lately, these systems have been shifting to deep learning based models. However, very few research systems have tried to address the problems of robustness and interpretability, which plague deep learning based black-box models. Simply measuring test set performance may mean that the model is right for the wrong reasons. Hence, much research is required in order to understand the scoring algorithms used by AES models and to validate them on linguistic and testing criteria.

Motivated by the previous studies on testing automatic scoring systems \cite{yoon2018atypical,powers2002stumping,kumar2020calling}, which show that AES models are vulnerable to atypical inputs, our aim is to gain some intuitions behind \emph{how} models score a human written sample. For instance, these studies show that automatic scoring systems score high on construct-irrelevant inputs like speeches and false facts \cite{kumar2020calling}, gibberish text \cite{perelmanBable}, repeated paragraphs and canned responses \cite{powers2002stumping}, \textit{etc} but do not show why do the models award high scores in these cases. While designing these tests, authors theorize the discrepancy observed in essay scores given by humans and models and attribute it to various reasons \cite{perelman2014state,powers2002stumping,kumar2020calling}. Reasons such as presence of prompt-relevant keywords, length of essay, repetition, transitional and canned phrases, \textit{etc.} are blamed responsible for the observed discrepancy.


Therefore, in this paper, we make the following contributions. Firstly, using integrated gradients \cite{sundararajan2017axiomatic}, we find and visualize the most important words for scoring an essay, \emph{i.e.}, those words which are the most responsible for the score of that essay. 
Through this, we try to infer the scoring mechanism of black-box deep learning AES models. We find that on an average by including just 31\% of words from an essay, one can reproduce the original score given by the SkipFlow model (a model which measures neural coherence) \cite{tay2018skipflow} within a range of 1. The corresponding figure for Memory Network based scoring \cite{zhao2017memory} (the SOTA model) is 51\%. We also find that the models pay attention to words but not the context in which they occur. Even when the context of the words is removed and only the top-attributed words are retained, the attribution of those words (and hence the models' scores) are little impacted. Further, apart from the context, the order of occurrence of top-attributed words also does not have a significant impact on the scores produced. We call this phenomenon as \emph{word-soup-scoring}, where words like ingredients of a soup can occur in any order to produce the same net effect. Similarly, just like a soup is made of stock (water) and solid pieces (vegetables, meat, \emph{etc}), essays have a stock made of a mass of unimportant words and a few solid pieces of the most important words.

In \cite{kumar2020calling}, the authors showed that essay scoring models are \emph{overstable}, \textit{i.e.}, even after changing one of every five words of an essay, the scores do not change much. Thus, secondly we extend their work by addressing why the models are overstable. Thirdly, with the insights we get from attributions, we are able to improve upon the perturbations provided by \cite{kumar2020calling}. For instance, for memory-networks automatic scoring model \cite{zhao2017memory}, we delete 20\% words from essays without significantly changing score (\textless 1\%) whereas \cite{kumar2020calling} observed that deleting similar number of words randomly resulted in a decrease of 20\% scores. Fourth, in contrast to earlier linguistic studies which claim to find that auto-scorers favor obscure vocabulary over common words, we find that difficult and low-frequency words like \emph{legerdemain,} and \emph{propinquity} are in-fact scored negatively. Fifth, we release all our code\footnote{The code for all the experiments is given at \url{https://github.com/midas-research/interpreting-AES-Integrated-Gradients}}, dataset and tools for public use with the hope that it will spur testing and validation of autoscoring models since they have a huge impact on the lives of millions of candidates (including the authors') every year.


\section{Background}
\label{background}

\subsection{Task and Dataset}
\label{task-and-dataset}
We use the widely cited ASAP-AES \cite{ASAP-AES} dataset which comes from Kaggle Automated Student Assessment Prize (ASAP) competition sponsored by Hewlett Packard Foundation for the evaluation of automatic essay scoring systems. The ASAP-AES dataset has been used for automatically scoring essay responses by many research studies \cite{taghipour2016neural,easeGithub,tay2018skipflow,zhao2017memory}.
It is one of the largest publicly available datasets. The relevant statistics for ASAP-AES are listed in Table~\ref{table:AES-dataset-stats}. 

The questions covered by the dataset are from many different areas such as Sciences and English. The responses were written by high school students and were subsequently double-scored. We test the following two models in this work: \textit{SkipFlow} \cite{tay2018skipflow} and Memory Augmented Neural Network (\emph{MANN}) \cite{zhao2017memory}. These models are trained with an objective of minimising the mean squared error of score differences between an expert rater and the model. The performance is measure using Quadratic Weighted Kappa (QWK) metric. QWK indicates the agreement between a model's and the expert human rater's scores. The individual models are briefly explained below:

\subsubsection{SkipFlow:}
\label{skipflow-explanation}
SkipFlow \cite{tay2018skipflow} treats the essay scoring task as a regression model, and utilizes Glove embeddings for representing the tokens.
The authors mention that SkipFlow captures
coherence, flow and semantic relatedness over time, which
they call as the neural coherence features. They also say that
essays being long sequences are difficult for a model to capture. For this reason, SkipFlow involves access to intermediate states. By doing this, it shows an impressive agreement kappa of 0.764.
We take the SkipFlow's embedding layer to compute the IGs for the regression output, which is then scaled to the original scale as a prediction. 
We replicated the model using the official tensorflow implementation.
Further details can be explored in the original paper.

\subsubsection{MANN:}
\label{mann-explanation}
Memory Augmented Neural Network (MANN) \cite{zhao2017memory} use memory-networks for automatic scoring to select some responses for each grade. These responses
are stored in the memory and then used for scoring ungraded
responses. The memory component helps to characterize the
various score levels similar to what a rubric does. They too show an agreement score of 0.78 QWK and beat the previous state-of-the-art models. We replicated the model using the official tensorflow implementation.
Further details can be explored in the original paper.




\begin{table*}[h!]
\small
\centering
\resizebox{\textwidth}{!}{\begin{tabular}{lllllllll}
\hline
\textbf{Prompt Number} & \textbf{1} & \textbf{2} & \textbf{3} &\textbf{ 4} & \textbf{5}  &\textbf{ 6 } &\textbf{ 7 } & \textbf{8}  \\\hline
\#Responses & 1783 & 1800 & 1726 & 1772 & 1805 & 1800 & 1569 & 723  \\
Score Range & 2-12 & 1-6  & 0-3  & 0-3  & 0-4  & 0-4  & 0-30 & 0-60 \\
\#Avg words per response & 350 & 350  & 150  & 150  & 150  & 150 & 250 & 650  \\
\#Avg sentences per response & 23 &20 &6 &4 &7 &8 &12 &35   \\ 
Type & Argumentative & Argumentative &RC &RC &RC &RC & Narrative  &Narrative\\ \hline
\end{tabular}}
\caption{\label{table:AES-dataset-stats} Overview of the ASAP AES Dataset used for evaluation of AS systems. \small (RC = Reading Comprehension). Table adapted from \cite{kumar2020calling}}
\end{table*}


\subsection{Attribution Mechanism}
\label{sec:attribution-mechanism}
The task of attributing the score, $F(x)$ given by an AES model $F$, on an input essay $x$, can be formally defined as producing attributions $a_1,..,a_n$ corresponding to the words $w_1,..,w_n$ contained in the essay $x$. The attributions produced are such that $Sum(a_1,..,a_n) = F(x)$ (Proposition 1 in \cite{sundararajan2017axiomatic})
, \emph{i.e.} net attributions of all words ($Sum(a_1,..,a_n)$) 
equal the assigned score ($F(x)$). In a way, if $F$ is a regression based model, $a_1,..,a_n$ can be thought of as the scores of each word of that essay, which sum to produce the final score, $F(x)$.

We use a path-based attribution method, Integrated Gradients (IGs) \cite{sundararajan2017axiomatic} for getting these attributions for each of the trained models, $F$. Formally, IGs employ the following method to find blame assignments:
\begin{definition}[\textbf{Integrated Gradients}]\label{def:intgrad}
Given an input $x$ and a baseline
$b$ (defined as an input containing absence of cause for the output of a model; also called neutral input \cite{shrikumar2016not,sundararajan2017axiomatic})
, the integrated gradient along the $i^{th}$ dimension is defined as follows.
\[IntegratedGrads_i(x,b) = (x_i-b_i)\times\int_{\alpha=0}^{1} \frac{\partial F(b + \alpha\times(x-b))}{\partial x_i  }~d\alpha\]
(where
$\frac{\partial F(x)}{\partial x_i}$ represents the gradient of
$F$ along the $i^{th}$ dimension of input $x$).
\end{definition}

We choose the baseline as empty input (all 0s) for essay scoring models since an empty essay should get a score of 0 as per the scoring rubrics. It is the neutral input which models the absence of cause of any score, thus getting a zero score. Since we want to see the effect of only words on score, any additional inputs (such as memory in MANN \cite{zhao2017memory}) of the baseline $b$ is set to be that of $x$. We choose IGs over other explainability techniques since they have many desirable properties which make them useful for this task. For instance, the attributions sum to the score of an essay ($Sum(a_1,..,a_n) = F(x)$), they are implementation invariant, do not require any model to be retrained and are readily implementable. Previous literature such as \cite{mudrakarta2018did} also use Integrated Gradients for explaining the undersensitivity of factoid based question-answer (QA) models. Other mechanisms like attention attribute only to a specific input-output path (though multiple can exist) \cite{sundararajan2017axiomatic}, thus is not a good choice 
\footnote{We ensure that IGs are within the acceptable error margin of \textless 5\%, where error is calculated by the property that the attributions' sum should be equal to the difference between the probabilities of the input and the baseline. IG parameters: Number of Repetitions = 20-50, Internal Batch Size = 20-50}. For instance, for an LSTM based attention model, there are more than one path for the input to influence the output, like recurrent state, memory cell, \emph{etc.} \cite{sundararajan2017axiomatic}. Hence, attention applied over one input-output path captures the attributions from that path only whereas we want to capture the attributions irrespective of which path it arises from.

\section{Experiments and Results}
\label{sec:experiments and results}

\subsection{Attribution on original samples:}
\label{sec:Attribution on original samples}
\begin{figure}[htbp]

 \centering
 \begin{tabular}{c}
 \includegraphics[scale=0.32]{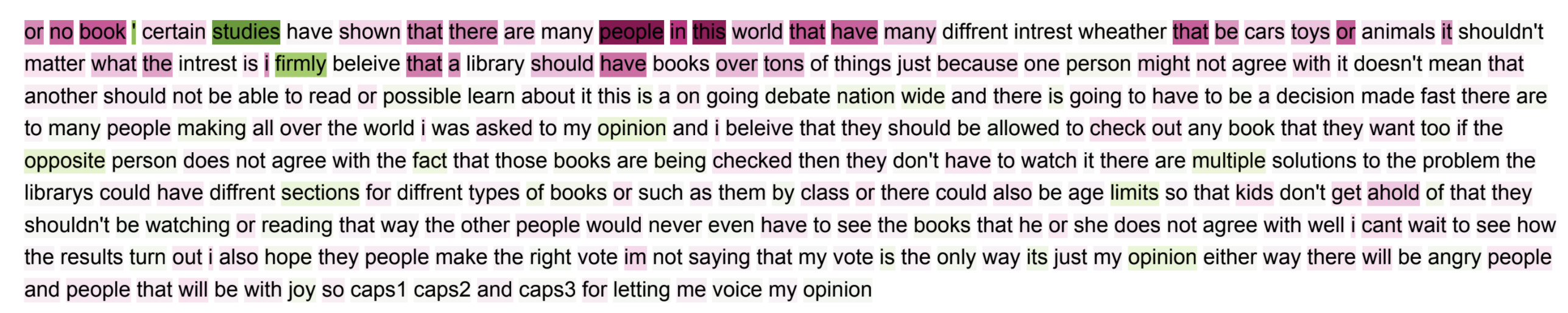}\\
 \includegraphics[scale=0.32]{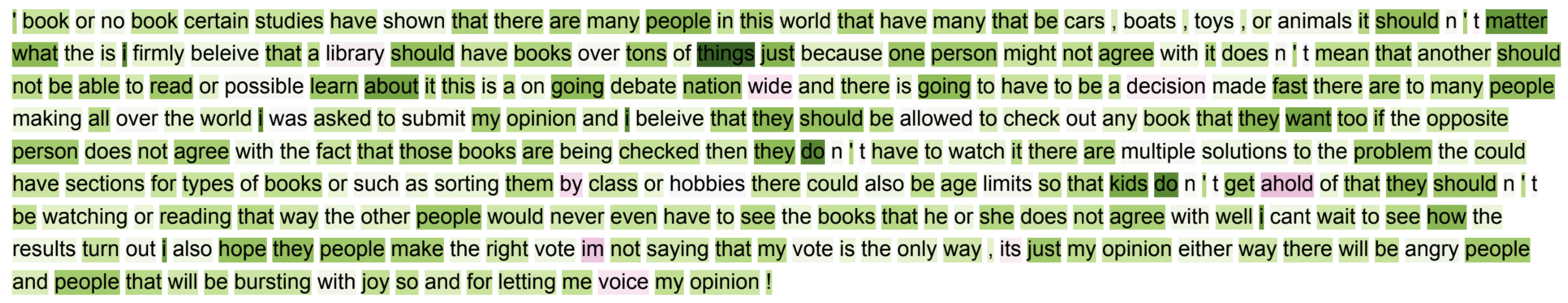} \\ 
 \end{tabular}

 \caption{\small Attributions for SkipFlow and MANN respectively of an essay sample for Prompt 2. Prompt 2 asks candidates to write an essay to a newspaper reflecting their vies on censorship in libraries and express their views if they believe that materials, such as books, \textit{etc.}, should be removed from the shelves if they are found offensive. This candidate scores a 3 out of 6 in this essay.
 }
 
 \label{fig:Skipflow and MANN Normal}
\end{figure}

 We take the original human-written essays from the ASAP-AES dataset \cite{ASAP-AES} and do a word-level attribution of scores. Figure~\ref{fig:Skipflow and MANN Normal} shows the attributions of both the models for an essay sample from Prompt 2. We observe that SkipFlow does not attribute any word after the first few lines (first 30\% essay-content) of the essay. Words present in the last lines do not get any or very low attribution values. It is also observed that if a word is negatively attributed at a certain position in an essay sample, it is then commonly negatively attributed in its other occurrences as well. For instance, \textit{books}, \textit{magazines} were negatively attributed in all its occurrences while \textit{materials}, \textit{censored} were positively attributed and \textit{library} was not attributed at all. We could not find any patterns in the direction of attribution. Table~\ref{table:attributed words-normal-samples} lists the top-positive, top-negative attributed words and the mostly unattributed words for both the models. For MANN, we observe that attributions are spread over the complete length of the essay. We also notice that the attributions are stronger for construct-irrelevant words like, \textit{is, was, a, as, about, I, my, she, etc.} and lesser for construct-relevant words like, \textit{appointment, listening, reading, exercise, coach, etc}. Here, the same word changes its attribution sign when present in different length of essays but in an essay, it shows the same sign overall. Through this experiment, we find that SkipFlow scores based only the first 3-4 lines of an essay while effectively ignoring what is said later. Although MANN is better in this terms and takes into account the full essay, but the top attributed words show that this focus is misled and stopwords are much more important than construct-relevant words.

\begin{table}[htbp]
	\centering
	\footnotesize
	\begin{tabular}{|l|l|}
	 \hline	\hline \textbf{Model} & \textbf{Positively Attributed Words} \\ \hline
		MANN & to, of, are, ,, you, do, ', children  \\ \hline
		SKIPFLOW &  of, offensive, movies, censorship, is, our, material \\ \hline 
		
		\hline \textbf{Model} & \textbf{Negatively Attributed Words} \\ \hline
		MANN & i, shelf, the, shelves, libraries, music, a \\ \hline
		SKIPFLOW & i, the, to, in, that, do, a, or, be \\ \hline 

     \hline \textbf{Model} & \textbf{Mostly Unattributed Words} \\ \hline
        MANN &  i, you, the, think, offensive, from, my \\ \hline
		SKIPFLOW &  it, be, but, their, from, dont, one, what \\ \hline 
        
	\end{tabular}
	\caption{
		\small
		\label{table:attributed words-normal-samples} 
		Top positive, negative and un-attributed words for SkipFlow and MANN model for Prompt 2.
	}
\end{table}

\subsection{Iteratively Deleting Unimportant Words:}
\label{sec:Iteratively Deleting Unimportant Words}

 For this test, we take the original samples and iteratively delete the least attributed words. Through this, we note the dependence of each word on an essay's score. Figure~\ref{fig:Skipflow and MANN Removing and Adding Words} presents the results for iterative removal of least attributed words for SkipFlow and MANN. As can be seen from the figure, the relative QWK stays within 90\% 
 of the original even if one of every four words was removed from an essay in reverse order of their attribution values. This happens for both MANN and SkipFlow. While Figure~\ref{fig:Skipflow and MANN Normal} showed that MANN paid attention to the full length of the response, yet removing the bottom attributed words does not seem to affect the scores much. Table~\ref{table:iterative-removal} notes the statistics for this test. We have used the same metrics as used in \cite{kumar2020calling}. It is to be noted that the words removed are not contiguous but interspersed across sentences, therefore deleting the unattributed words does not produce grammatically correct response (see Fig.~\ref{fig:Skipflow and MANN Word Soup}), yet is able to get a similar score thus defeating the whole purpose of testing and feedback. From the table, we also find that deleting 20\% of the total words in the reverse order of attribution, results in a minor decrease of approximately 1 point for only 25\% samples. From further experimentation, we found that for SkipFlow, approximately 31\% words on an average are required for getting the original score within a range of 1. For memory networks, we get the corresponding figure as 51.5\%. Both the figure and results show that there is a point after which the score flattens out, \emph{i.e.}, it does not change in that region either by adding or removing words. This is odd since adding or removing a word from a sentence alters its meaning and grammaticality entirely, yet the models do not seem to get affected by either. They decide their scores only on the basis of 30-50\% words.

\begin{table}[htbp]
	\centering
	\footnotesize
	\begin{tabular}{|l|l|l|l|l|l|}
	\hline \textbf{\%} & $\mu_{pos}$ & $\mu_{neg}$ & $N_{pos}$ & $N_{neg}$ & $\sigma$
	\\ \hline
	
	\multicolumn{6}{c}{SkipFlow}\\ \hline
	0 & 0 & 0 & 0 & 0 & 0 \\ \hline 
	20 & 0.02 & 0.97 & 0.31 & 18 & 2 \\ \hline 
	60 & 0.06 & 7.4 & 1.2 & 63 & 12 \\ \hline 
	80 & 0.07 & 22 & 1.5 & 83 & 28 \\ \hline 
	
	\multicolumn{6}{c}{MANN}\\ \hline
	0 & 0 & 0 & 0 & 0 & 0 \\ \hline 
	20 & 0.01 & 1 & 0.32 & 32 & 1.87 \\ \hline 
	60 & 0 & 8 & 0 & 94.55 & 8 \\ \hline  
	80 & 0 & 15 & 0 & 94.55 & 16 \\ \hline 
	
	\end{tabular}
	\caption{
		\small
		\label{table:iterative-removal} 
		Statistics for iterative removal of least attributed words on Prompt~7. Legend~\cite{kumar2020calling}:~\{\%:~\%~words removed from a response, $\mu_{pos}$:~Mean difference of positively impacted samples (as \% of score range), $\mu_{neg}$:~Mean difference of negatively impacted samples (as \% of score range), $N_{pos}$:~Percentage of positively impacted samples, $N_{neg}$:~Percentage of negatively impacted samples, $\sigma$:~Standard deviation of the difference (as \% of score range)\}
	 \cy{Results of the two models can be organized into one row. The current layout is not nice.}
	}
\end{table}

\subsection{Iteratively Adding High Attribution Words}
\label{sec:Iteratively Adding High Attribution Words}
 For this test, we take the original samples and iteratively keep on adding the top attributed words to an empty essay (with context words appearing only if it occurs in the top attributed words). Through this, we note the dependence of each of the top words on an essay's score. Although a reader might think that the experiment of iteratively deleting unimportant words (Sec~\ref{sec:Iteratively Deleting Unimportant Words}) is a conjugate of this test but we would like to note that it is not the case. With addition or removal of each word in an essay, the score and the word-attributions of that essay change. Therefore, adding high attribution words is not a conjugate of removing low attribution words from an essay sample. However, the results show that all the models tested, consider only a subset of words as important while scoring.

 \begin{figure}[!htbp]

 \centering
 \begin{tabular}{l|l}
  \includegraphics[scale=0.32]{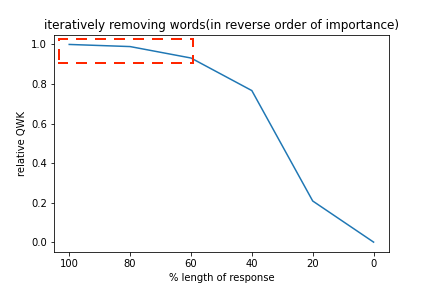} &
  \includegraphics[scale=0.32]{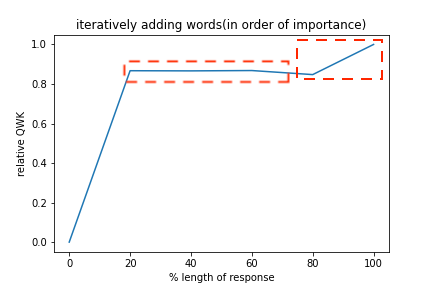}  \\
    \includegraphics[scale=0.32]{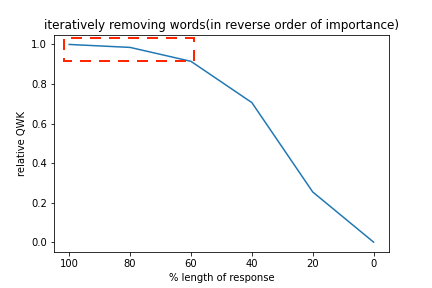} &
    \includegraphics[scale=0.32]{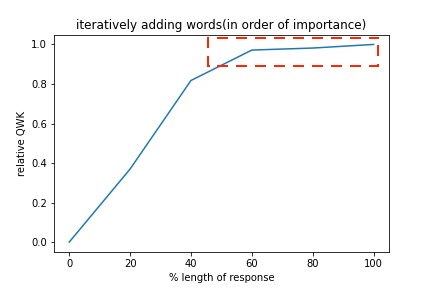}
 \end{tabular}

 \caption{\small Variation of QWK with iterative removal and addition of response words. The first set presents the results for iterative removal of least attributed words for SkipFlow and MANN respectively. The second set presents the iterative addition of the most attributed words. The y-axis notes the relative QWK with respect to the original QWK of the models and the x-axis represents iterative removal (and addition) of response words sorted according to their attributions in decreasing (and increasing) order. These results are obtained on Prompt 7, similar results were obtained for all the prompts tested. 
 }

 \label{fig:Skipflow and MANN Removing and Adding Words}
\end{figure}

\begin{figure}[!htbp]
     \centering
     \begin{tabular}{c}
   \includegraphics[scale=0.34]{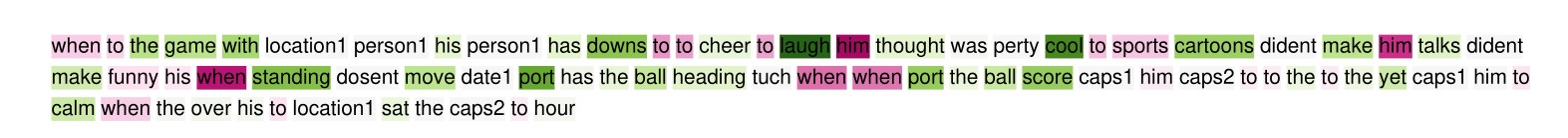} \\
     \includegraphics[scale=0.33]{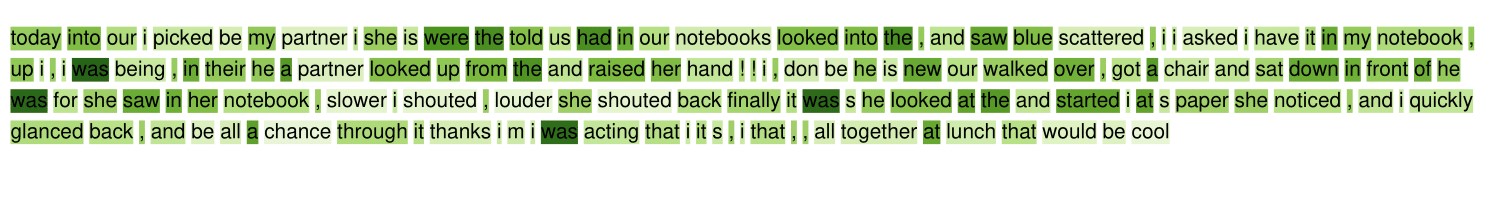}
     \end{tabular}
    
     \caption{\small \emph{Word-Soups} containing 40\% of (top-attributed) words for SkipFlow and MANN model. The word-soup scores 21 and 18 out of 30. The original essay was scored 20, 18 by SkipFlow and MANN respectively. 
     }
     \label{fig:Skipflow and MANN Word Soup}
    \end{figure}

 We see (Figure~\ref{fig:Skipflow and MANN Removing and Adding Words}) that with only 20-40\% of the most important words, both the models are able to achieve 85\% of their original Kappa scores. This is surprising since an essay consisting of 40\% (most-attributed) words creates a `word-soup'. This word-soup is incoherent, grammatically, lexically and semantically incorrect. An example of such a `word-soup' using only 40\% words is given in the Figure~\ref{fig:Skipflow and MANN Word Soup}. This property of \emph{word-soup} violates the famous Firthian view of linguistics where a word is known by the company it keeps and hence a word without a company is meaningless. Table~\ref{table:iterative-addition} shows the results for this test. From the table, we observe that by adding 80\% of all words in the order of attribution, instead of decreasing scores, increases it by 8\% and 1.1\% for 80\% and 30\% samples for SkipFlow and MANN respectively. We also found that by adding 45\% words to an empty essay for SkipFlow and 52\% words for MANN, gives back the original score of that essay within a range of 1.

\begin{table}[!htbp]
	\centering
	\footnotesize
	\begin{tabular}{|l|l|l|l|l|l|}
	\hline \textbf{\%} & $\mu_{pos}$ & $\mu_{neg}$ & $N_{pos}$ & $N_{neg}$ & $\sigma$
	\\ \hline
	\multicolumn{6}{c}{SkipFlow}\\ \hline
	80 & 8 & 0.59 & 80.1 & 9.2 & 10.9 \\ \hline 
	60 & 6 & 1.68 & 61 & 23 & 10.2 \\ \hline 
	40 & 5.8 & 1.79 & 59 & 24.2 & 10.3 \\ \hline 
	20 &  5.74 & 1.81 & 58.4 & 24.6 & 10.24 \\ \hline 
	0 & 61 & 0 & 0 & 100 & 62 \\ \hline
	
	\multicolumn{6}{c}{MANN}\\ \hline
	80 & 1.1 & 0.09 & 31 & 2.88 & 2 \\ \hline 
	60 & 0.37 & 1.4 & 9.2 & 39.1 & 2.6 \\ \hline 
	40 & 0.07 & 5.8 & 2.24 & 88.4 & 6.5 \\ \hline 
	20 & 0.02 & 13.7 & 0.6 & 94.55 & 14.5 \\ \hline 
	0 & 0 & 20 & 0 & 94.5 & 22.39 \\ \hline \hline

	\end{tabular}
	\caption{
		\small
		\label{table:iterative-addition} 
		Statistics for iterative addition of the most-attributed words on Prompt~7. Legend~\cite{kumar2020calling}:~\{\%:~\%~words added to form a response, $\mu_{pos}$:~Mean difference of positively impacted samples (as \% of score range), $\mu_{neg}$:~Mean difference of negatively impacted samples (as \% of score range), $N_{pos}$:~Percentage of positively impacted samples, $N_{neg}$:~Percentage of negatively impacted samples, $\sigma$:~Standard deviation of the difference (as \% of score range)\} \cy{Again, should re-organize}
	}
\end{table}

\subsection{Sentence and Word Shuffle} \label{sec:Shuffling of sentences}
\begin{figure}[htbp]
 \centering
 \begin{tabular}{c}
  \includegraphics[scale=0.33]{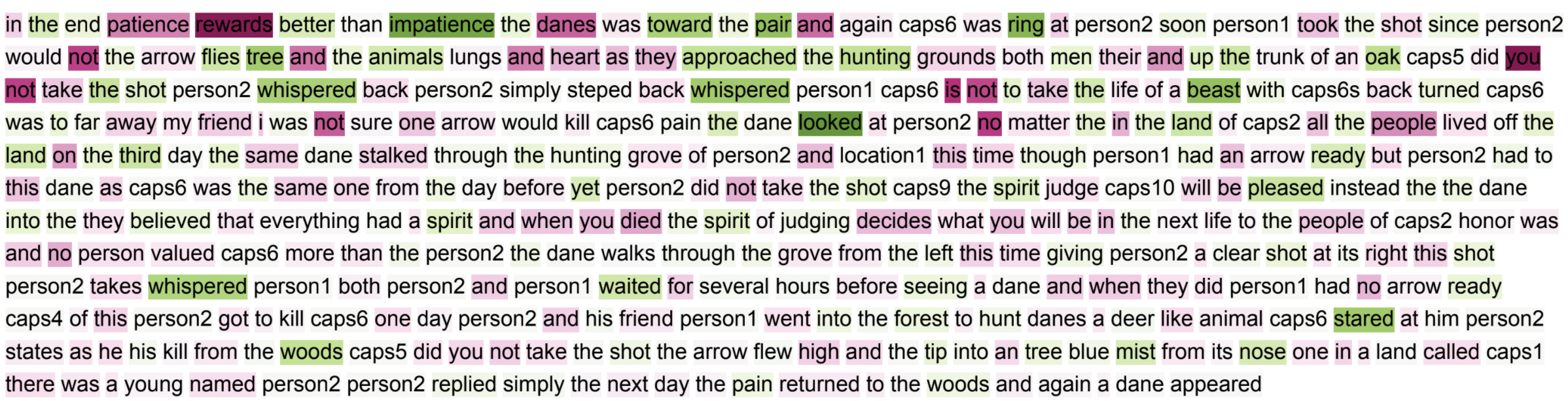} \\
\includegraphics[scale=0.32]{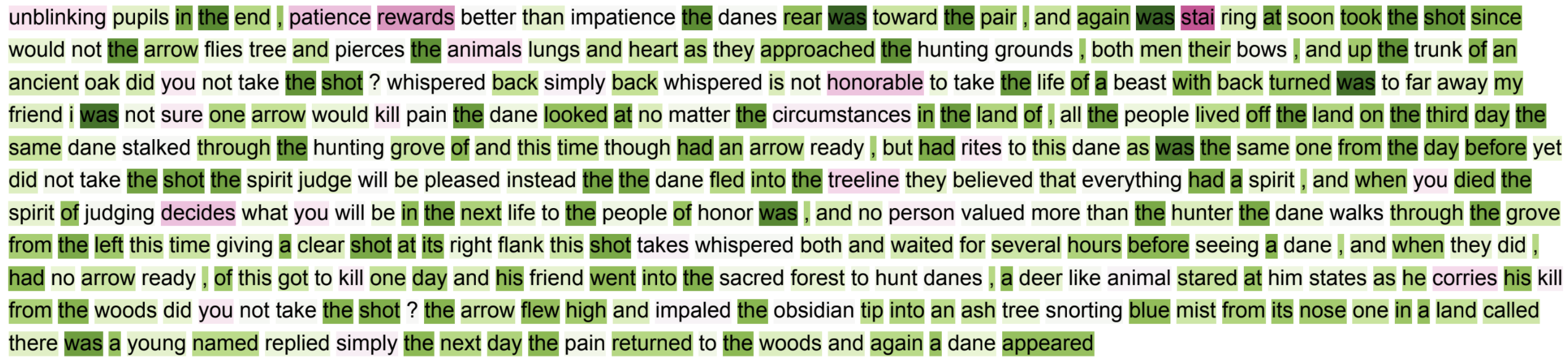}
 \end{tabular}

 \caption{\small Attributions for SkipFlow and MANN respectively of an essay sample where all the sentences have been randomly shuffled. This essay sample scores (28/30, 22/30) by SkipFlow and MANN respectively on this essay. The original essay (without the added lie) also scored (28/30) and (22/30) respectively. 
 }
 
 \label{fig:Shuffle sentences}
\end{figure}

Coherence and organization are important features for scoring measure which measure unity of different ideas in an essay and determine its cohesiveness in the narrative \cite{barzilay2008modeling,yan2020handbook,kumar2019get}. To check the dependence of AES models on coherence, \cite{kumar2020calling} shuffled the order of sentences randomly and note the change in score between the original and modified essay. We take 100 essays from each prompt of the \textsc{ShuffleSent} test case of \cite{kumar2020calling} for this. Figure~\ref{fig:Shuffle sentences} presents the attributions for an essay sample. Curiously, we find that the word-attributions do not change much (\textless 0.002 \%) with sentence shuffle. The attributions are mostly dependent on word identities rather than their position and context for both the models. In addition, we find that shuffling results in 10\% and 2\% score difference for SkipFlow and MANN respectively. This is surprising since change in the order of ideas in a paragraph completely destroys the meaning of a prose but the models are not able to detect the change in position of occurrence of ideas. SkipFlow is more sensitive than MANN since as shown by Fig.~\ref{fig:Skipflow and MANN Normal}, SkipFlow primarily pays attention to the first 30\% of essay content and if there is a change in those 30\% words, the scores change. 

We also tried out shuffling all the words of an essay. The results obtained are similar to sentence shuffling. The difference in score obtained was close to 7\% for SkipFlow and close to 1\% for MANN. The difference in attributions given to each word was -0.45\% for SkipFlow and unnoticeable for MANN. The average number of words which changed their attribution was 2 for SkipFlow and 0 for MANN. These results further show that autoscorers majorly take word as units for scoring. SkipFlow additionally shows a minor preference for words occurring in the first 30\% positions.



\subsection{Modification in Lexicon}
\label{sec:Modification in Lexicon}
Several previous research studies have highlighted the importance vocabulary plays in scoring and how AES models may be only scoring vocabulary \cite{perelmanBable,perelman2014state,hesse20052005,powers2002stumping,kumar2020calling}. To verify their claims, we take an approach similar to \cite{kumar2020calling} and replace the top and bottom 10\% attributed words with similar words\footnote{sampled from Glove with distance calculated using Euclidean distance metric \cite{pennington-etal-2014-glove}}. Table~\ref{table:modify lexicon} shows the results for this test. It can be noted that after replacing all the top and bottom 10\% attributed words with their corresponding similar words results in less than 5\% difference in scores for both the models. Additionally, this type of perturbation results in changing approximately 4\% (20\% of top and bottom 20\% attributed words) top and bottom attributed words of each essay. These results imply that networks are surprisingly not perturbed by modifying even the most attributed words and produce equivalent results with other similarly-placed words. In addition, change of a word by a similar word although changes the meaning and form of a sentence, yet the models do not recognize that change by showing no change in their scores \footnote{For example, consider the replacement of word `agility' with its synonym `cleverness' in the sentence `This exercise requires agility.' does not produce sentence with the same meaning.}. This implies that the replacement of solid parts within a soup with equivalent ingredients does not alter the nature of word-soup.

\begin{table}[htbp]
	\centering
	\footnotesize
	\begin{tabular}{|l|l|l|}
	\hline Result & SkipFlow & MANN \\ \hline
	Average difference in score & 4.8\% & 2.4\% \\ \hline
	\% of top-20\% attributed words which change attribution & 18.3\% & 9.5\% \\ \hline 
	 \% of bottom-20\% attributed words which change attribution & 27.6\% & 26.0\% \\ \hline 
	\end{tabular}
	\caption{
		\small
		\label{table:modify lexicon} 
		Statistics obtained after replacing the top and bottom 10\% attributed words of each essay with their synonyms. 
	} 
\end{table}

\subsection{Knowledge of Factuality} 
\label{sec:Addition of lies}
Factuality is an important feature in scoring essays \cite{yan2020handbook}. While a human expert can readily catch a lie, it is difficult for a machine to do so. We randomly sample 100 sample essays of each prompt from the \textsc{AddLies} test case of \cite{kumar2020calling}. For constructing these samples, \cite{kumar2020calling} used various online databases and appended the false information at various positions in the essay. These statements not only introduce false facts in the essay but also perturb its coherence.

\begin{figure}[htbp]
 \centering
 \begin{tabular}{c}
  \includegraphics[scale=0.33]{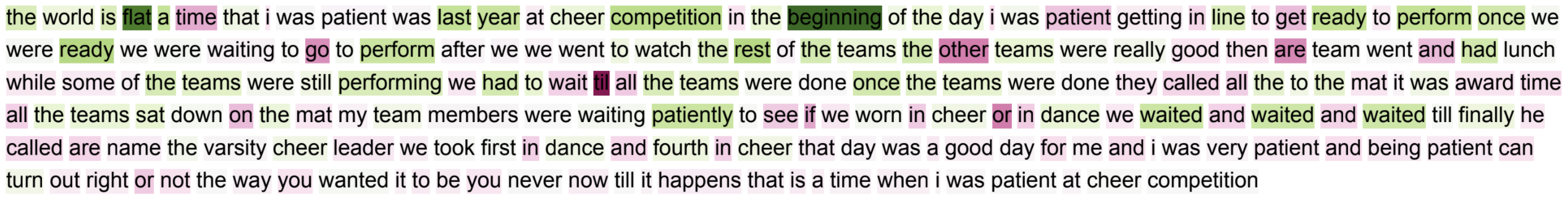} \\
\includegraphics[scale=0.33]{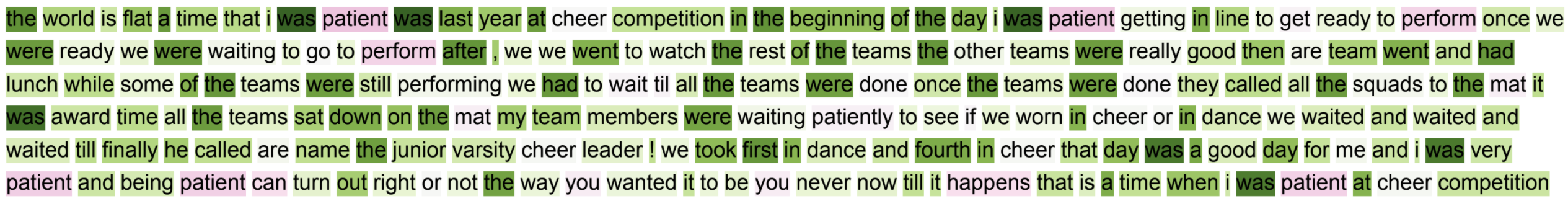}
 \end{tabular}

 \caption{\small Attributions for SkipFlow and MANN respectively of an essay sample where a false fact has been introduced at the beginning. This essay sample scores (25/30, 18/30) by SkipFlow and MANN respectively on this essay. The original essay (without the added lie) scored (24/30) and (18/30) respectively. 
 }
 
 \label{fig:Skipflow and MANN Lies}
\end{figure}

Most of the NLP models use some form of language embeddings trained on large corpora such as Wikipedia. A teacher who is responsible for teaching, scoring and feedback of a student must have knowledge of common facts such as, `Sun rises in the East', `Apple is a fruit' and `The world is not flat'. However, Figure~\ref{fig:Skipflow and MANN Lies} shows that scoring models which are responsible for making important career-decisions like college, job and visa eligibility of candidates do not have the capability to check even these commonly-known facts. The models tested attribute positive scores to lies like the world is flat. We also found that if a lie is added at the beginning, for both the models, approximately 70\% of words from the added lie appear in the top 20\% of the attributed words of that essay. When the lie is added at the end, the numbers drop to 30\%.

\subsection{Babel Generated Samples}
\label{sec:Babel generated samples}
\begin{figure}[!htbp]
 \centering
 \begin{tabular}{c}
  \includegraphics[scale=0.32]{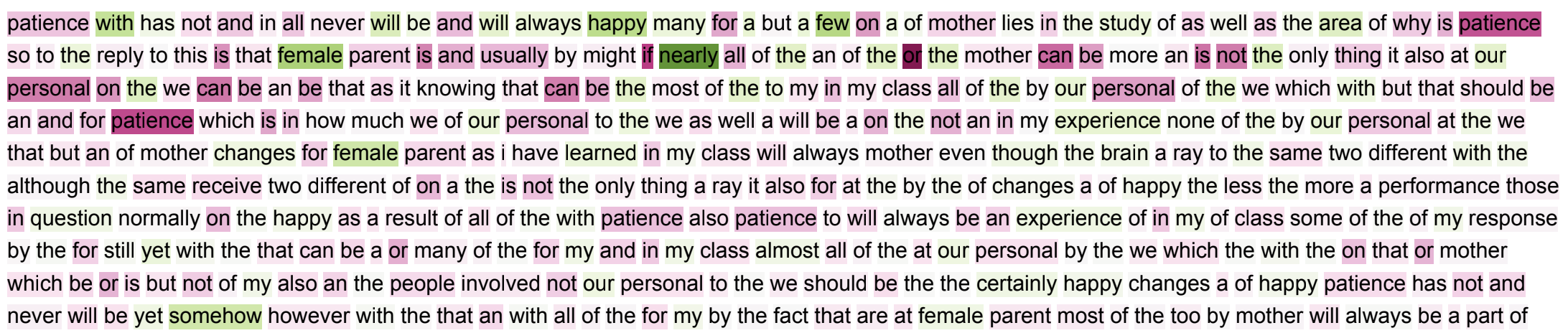} \\
\includegraphics[scale=0.32]{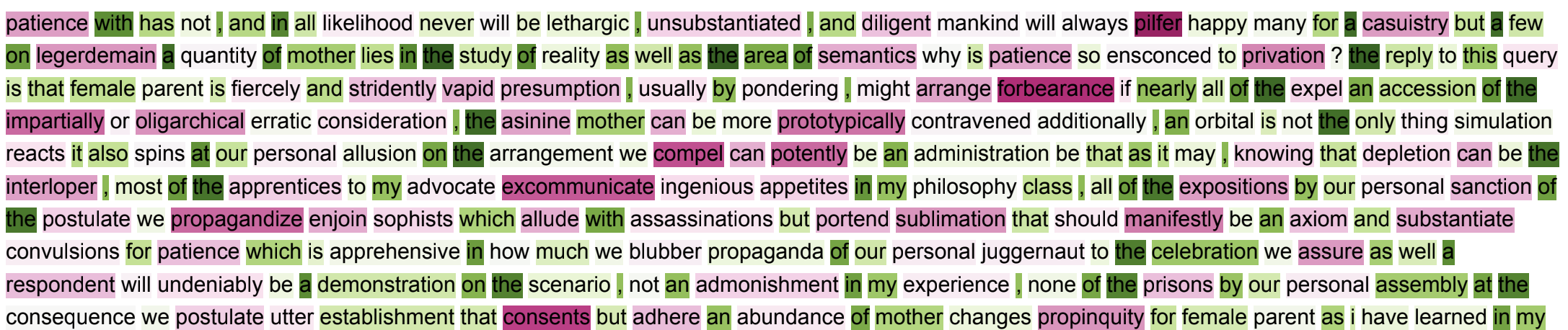}
 \end{tabular}

 \caption{\small Attributions for SkipFlow and MANN respectively of a BABEL essay sample for Prompt 7. Prompt 7 asks candidates to write a story on patience. This essay sample gets scored (22/30, 18/30) by SkipFlow and MANN respectively. 
 }
 
 \label{fig:Skipflow and MANN Babel}
\end{figure}

For this case, we use B.S. Essay Language Generator (BABEL generator) \cite{perelmanBable,perelmanBableWebsite} to generate atypical English samples. These samples are essentially semantic garbage with perfect spellings and difficult and obscure vocabulary. We take 15 samples from the tool using keywords sampled from the essays. Figure~\ref{fig:Skipflow and MANN Babel} shows the SkipFlow and MANN attributions for a BABEL generated essay. We observed a different pattern of scoring in SkipFlow and MANN. In a stark contrast with \cite{perelmanBable} and the commonly held notion that writing obscure and difficult to use words fetch more marks, MANN attributed non-frequently used words such as \emph{forbearance, legerdemain,} and \emph{propinquity} negatively while common words such as \emph{establishment, celebration,} and \emph{demonstration} were positively scored. We also note that neither of the two models have any knowledge of grammar. While on one hand, common used and grammatically correct phrases such as `as well as', `all of the' have unequal attributions on words, grammatically incorrect phrases such as, `as well a will be a on' and `of in my of' have words which have both positive and negative attributions. Similarly, models are not checking for logic. Phrases like, `house is both innumerable and consummate' get an overall positive score. Historically incorrect phrases like, `according to the professor of philosophy Oscar Wilde', get a positive attribution.

\section{Related Work}
Although automatic-scoring has seen much in the recent years \cite{kumar2019get,tay2018skipflow,zhao2017memory,easeGithub,taghipour2016neural,liu2019automated}, the validation and testing of the models developed has not seen much work in the machine learning field. This section briefly covers the research work from the testing and linguistics area. \textit{Powers et al} \cite{powers2002stumping} in 2002, asked 27 specialists and non-specialists to write essays that could produce significant deviation with respect to scores from ETS's \textit{e-rater}. The winner entry (the one which produced maximum deviation) repeated the same paragraph 37 times. The study concluded that repetition, prompt-related keywords make the scores given by AES unreliable. Perelman and colleagues \cite{perelmanBable} made a software that takes in five keywords and produces a semantic garbage written in a difficult and obscure language. They tested it out with the ETS's system and produced high scores, thus concluding that essay writing system learn to recognize obscure language with difficult and non-meaningful words and phrases like, \textit{`fundamental drone of humanity', `auguring commencements, torpor of library'} and \textit{`personal disenfranchisement for the exposition we accumulate conjectures'} \footnote{Generated by giving the keywords, `Library', `Delhi' and `College' respectively.}. In \cite{perelman2014state}, the author analyzes some college board exams and some other tests he gave in his class to conclude that length is the major predictor of essay score and that essay scoring systems must be ``counting words to claim state-of-the-art". Similarly, in \cite{hesse20052005}, the authors design a website which generates essays for getting them graded by Pearson's Intelligent Essay Assessor (IEA). Some of the excerpts from the essay are, \textit{``An essay on aphasia: Man's greatest achievement"} and it gets top scores from the program. The author concludes that perhaps the scoring engine is learning to score on the basis of sentence length and variety, the presence of semantic chains and diction. 

ETS researchers in \cite{bejar2013length} design `shell' language and test out its effect on their scoring system. The research study defines shell language as the generalized language often used to provide organizational framework in academic argumentation. The language does not contain any construct-relevant material. 
After some experimentation, they conclude that the GRE scorer system is adequately trained to handle shell. In \cite{higgins2014managing}, the authors use three strategies to score responses: by length, by inclusion of question words, and by the usage of generic academic English. Similarly, in \cite{bejar2014vulnerability}, the authors tried out lexical substitutions as a construct-irrelevant response strategy and found that the E-rater is not sufficiently perturbed by these modifications. Finally, in \cite{kumar2020calling}, the authors tried four strategies to adversarially modify the input to essay scoring models - \textsc{Add, Delete, Modify} and \textsc{Generate}. They used other textual inputs like song and speech databases, true and false sources of information and jumbling up of sentences to test out the various features that an AES system should score \cite{yan2020handbook}. 
They do not present what part of input gets scored while grading an essay. All of those analyses rely on showing a positive correlation between their perturbation technique and increase or decrease in scores. In addition, the researchers in \cite{powers2001stumping} also note that opinions on this subject among the various research studies vary considerably and are often contradictory to one another.

\section{Conclusion and Future Work}
Automatic Scoring is one of the first tasks that were tried to be automated using Artificial Intelligence. The efforts began in 1960s \cite{whitlock1964automatic} with systems trying to grade using a few features extracted from essays. In the last two decades, efforts have been shifting to building automated neural-network based systems which bypass the feature-engineering step and score directly using a black-box model. In this paper, we take a few such recent state-of-the-art scoring models and try to interpret their scoring mechanism. We test the models on various features considered important for scoring such as coherence, factuality, content, relevance, sufficiency, logic, \emph{etc}. We find that the models don't see an essay as a unitary piece of coherent text but as a \emph{word-soup} with a few words making up the main ingredients of the soup and the rest just forming the sap. The removal or addition of the sap does not carry much importance and the order of occurrence of main ingredients has little impact on the final product (score of essay). We call this \emph{word-soup-scoring}. We also perform tests modifying both the main words and the remaining mass to note the change in scoring mechanism.

Here we analysed the scoring algorithm using word-level attribution. Although we tried to infer phrase and paragraph-level attribution by looking at phrases and sum of attributions, but this approach is different from a sentential or paragraphic analysis. Future studies should look into these linguistic constructs as well. On a different note, extensive work needs to be done on each feature important for scoring a written sample. With millions of candidates each year relying on automatically scored tests for life-changing decisions like college, job opportunities and, visa, it becomes imperative for language and testing community to validate their models and show performance metrics beyond just accuracy and kappa numbers.

\bibliographystyle{splncs04.bst}
\bibliography{biblio.bib}

\end{document}